\documentclass[runningheads]{llncs}
\usepackage{graphicx}
\usepackage{booktabs}
\usepackage{amsfonts}
\usepackage{multirow}
\usepackage{pifont}
\usepackage{array}
\usepackage{amsmath}
\usepackage{xcolor,colortbl}
\usepackage{nicematrix}

\usepackage{hyperref}
\definecolor{Gray}{gray}{0.9}
\definecolor{palegray}{HTML}{F5F5F5}
\definecolor{barred}{HTML}{F24B4B}
\definecolor{barblue}{HTML}{1DB6F2}
\definecolor{baryellow}{HTML}{F2A922}
\definecolor{barpurpule}{HTML}{A441BF}
\definecolor{bargreen}{HTML}{97BF5A}
\definecolor{palepurpule}{HTML}{FFF8FF}
\definecolor{paleblue}{HTML}{AFEEEE}
\definecolor{palegreen}{HTML}{F6FAF3}
\newcolumntype{u}{>{\columncolor{gray!15}}c}
\newcolumntype{g}{>{\columncolor{green!15}}c}
\newcolumntype{k}{>{\columncolor{blue!15}}c}
\newsavebox\CBox 
\def\textBF#1{\sbox\CBox{#1}\resizebox{\wd\CBox}{\ht\CBox}{\textbf{#1}}}
    \makeatletter
\def\@fnsymbol#1{\ensuremath{\ifcase#1\or *\or \dagger\or \ddagger\or
   \mathsection\or \mathparagraph\or \|\or **\or \dagger\dagger
   \or \ddagger\ddagger \else\@ctrerr\fi}}
    \makeatother

\makeatletter
\newcommand{\printfnsymbol}[1]{%
  \textsuperscript{\@fnsymbol{#1}}%
}
\makeatother

\begin{document}

\title{Open-Ended Medical Visual Question Answering Through Prefix Tuning of Language Models}
\titlerunning{Open-Ended Medical VQA Through Prefix Tuning of Language Models}
\author{Tom van Sonsbeek\thanks{Equal contribution}\thanks{Corresponding author} \and 
Mohammad Mahdi Derakhshani\printfnsymbol{1} \and
Ivona Najdenkoska\printfnsymbol{1} \and\\
Cees G. M. Snoek \and
Marcel Worring}

\authorrunning{T. van Sonsbeek et al.}
%
\institute{University of Amsterdam, Amsterdam, the Netherlands\\\email{\{t.j.vansonsbeek,m.m.derakhshani,i.najdenkoska,c.g.m.snoek,m.worring\}@uva.nl}}
\maketitle             
\begin{abstract}
Medical Visual Question Answering (VQA) is an important challenge, as it would lead to faster and more accurate diagnoses and treatment decisions. Most existing methods approach it as a multi-class classification problem, which restricts the outcome to a predefined closed-set of curated answers. We focus on open-ended VQA and motivated by the recent advances in language models consider it as a generative task. Leveraging pre-trained language models, we introduce a novel method  particularly suited for small, domain-specific, medical datasets. To properly communicate the medical images to the language model, we develop a network that maps the extracted visual features to a set of learnable tokens. Then, alongside the question, these learnable tokens directly prompt the language model. We explore recent parameter-efficient fine-tuning strategies for language models, which allow for resource- and data-efficient fine-tuning.  
We evaluate our approach on the prime medical VQA benchmarks, namely, Slake, OVQA and PathVQA. The results demonstrate that our approach outperforms existing methods across various training settings while also being computationally efficient.

\keywords{Visual Question Answering  \and Language Models \and Prompting \and Prefix Tuning.}
\end{abstract}
\section{Introduction}
Images and text are inherently intertwined in clinical diagnosis and treatment. Having an automated approach that is able to answer questions based on images, giving insight to clinicians and patients, can be a valuable asset. In such a medical Visual Question Answering (VQA) setting the common approach is to treat VQA as a multi-class classification problem solved by neural networks. Given a joint encoded representation of the image and question, the model classifies it into a predefined set of answers. Although these approaches yield good performance~\cite{do2021multiple,wu2022medical,lin2021medical,nguyen2019overcoming}, they deal with closed-set predictions, which is not an ideal solution for VQA. For instance, medical VQA datasets commonly contain hundreds to thousands of free-form answers~\cite{he2020pathvqa}, which is suboptimal to be treated as a classification task. Moreover, the severe class imbalance and out-of-vocabulary answers further hinder the generalizability of these classification methods.

We believe that a possible solution can be found in the generative capability of language models, since they are able to produce free text, instead of being limited to closed-set predictions. However, leveraging language models for solving open-ended medical VQA is limited due to several challenges, such as finding ways to properly communicate the visual features and letting such large-scale models be employed on small-sized medical VQA datasets.

Inspired by recent image captioning models~\cite{mokady2021clipcap}, we propose to use the medical images by converting them into a set of learnable tokens through a small-scale mapping network. These tokens can then be interpreted as a visual prefix for the language model~\cite{barraco2022unreasonable,najdenkoska2023meta,derakhshani2022variational}. Afterward, the visual prefix is used together with the question as input to the language model, which generates the answer token by token~\cite{radford2021learning}.

Furthermore, large-scale language models can generalize across domains while keeping their weights frozen~\cite{tsimpoukelli2021multimodal}. This makes them very appealing for the medical domain, which inherently does not possess large quantities of labeled data required to train these models from scratch~\cite{taylor2022clinical}.
Models like BioGPT~\cite{luo2022biogpt} and BioMedLM~\cite{biomedlm} are based on the generic GPT2 language model~\cite{radford2019language} and are trained on biomedical text corpora. They perform quite well compared to their general counterparts on specific biomedical language tasks, like question answering or relation extraction. We design our model in a flexible manner, which allows us to incorporate any of these pre-trained language models.

In summary,  we contribute in three major aspects: (i) We propose the first large-scale language model-based method for open-ended medical VQA. (ii) We adopt parameter-efficient tuning strategies for the language backbone, which gives us the ability to fine-tune a large model with a small dataset without the danger of overfitting. (iii) We demonstrate through extensive experiments on relevant benchmarks that our model yields strong open-ended VQA performance without the need for extensive computational resources.

\section{Related works}
To describe existing medical VQA methods, we make a distinction between classification methods and generative methods. The majority of methods are classification-based and make use of different types of encoders, such as CNNs or Transformers \cite{he2020pathvqa,wang2022mhkd,gong2021cross,cong2022caption,li2022bi} followed by a classification layer. 

\subsubsection{Classification-based VQA} We highlight a number of methods that showed good performance on current competitive medical VQA datasets. The Mixture Enhanced Visual Features (MEVF)~\cite{nguyen2019overcoming} is initialized based on pre-trained weights from the Model-Agnostic Meta-Learning (MAML) model \cite{finn2017model} in combination with image feature extraction from Conditional Denoising Auto-Encoders (CDAE) to generate a joint question-answer representation using Bilinear (BAN) or Stacked (SAN) Attention Networks. Do \textit{et al.}~\cite{do2021multiple} create a similar embedding space by extracting annotations from multiple pre-trained meta-models, and learning meta-annotations by training each meta-model. Linear combinations~\cite{gong2022vqamix} or question-conditioned selections~\cite{zhan2020medical} from this multi-modal embedding space can further enhance performance. The use of Transformer~\cite{khare2021mmbert} and CLIP~\cite{radford2021learning,eslami2021does} encoders also results in strong VQA classification performance. 

\subsubsection{Open-ended VQA} MedFuseNet~\cite{sharma2021medfusenet} is one of the few methods performing and reporting open-ended visual question answering on recent public datasets. They do so by creating a BERT-based multi-modal representation of image and question and subsequently passing it through an LSTM decoder. Ren \textit{et al.}~\cite{ren2020cgmvqa} create open-ended answers by using the masked token prediction functionality of BERT. We aim to show that generative language models are more versatile and better suited for this task. 
\section{Methodology}
\subsection{Problem Statement}
Given an input image $\mathbf{I}$ and an input question in natural language $\mathbf{Q}$, our method aims to sequentially generate an answer $\mathbf{A}=\{A_0, A_1, ..., A_N\}$ composed of $N$ tokens, by conditioning on both inputs. From a model definition perspective, we aim to find the optimal parameters $\theta^*$ for a model by maximizing the conditional log-likelihood as follows:
\begin{equation}
\theta^* = \underset{\theta}{\arg\max} \sum^N_{i=1}  \log p_{\theta}(\mathbf{A}_i|\mathbf{Q}, \mathbf{I},\mathbf{A}_{i-1}). 
\label{eq:1}
\end{equation}

\subsection{Model Architecture}
\begin{figure}[!t]
    \centering
    \includegraphics[width=\linewidth]{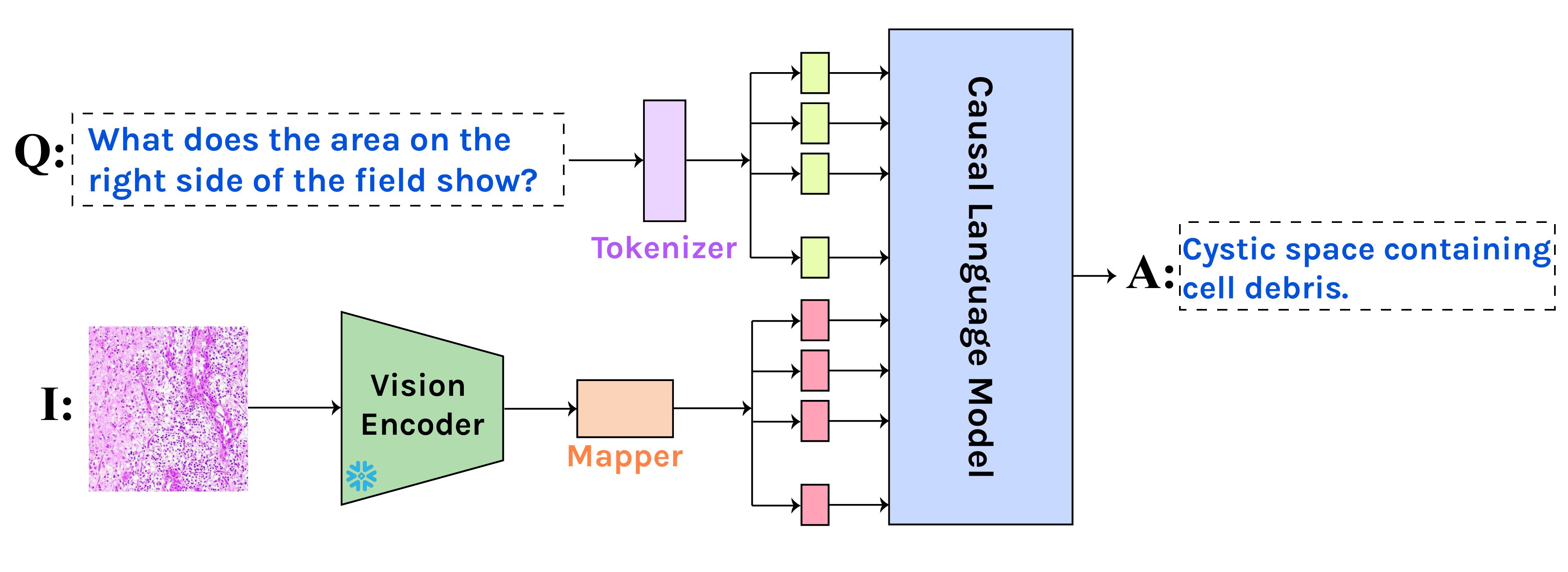}
    \caption{Model architecture of our proposed open-ended generative VQA method.}
    \label{fig:model_arch}
\end{figure}

Our VQA model is designed as an encoder-decoder architecture, with a two-stream encoder and a language model~(LM) as a decoder, as illustrated in Fig.~\ref{fig:model_arch}. Specifically, the two streams encode the two input modalities, namely the image $\mathbf{I}$ and the question $\mathbf{Q}$. The language model is defined as a causal language Transformer~\cite{radford2019language}, and it generates the answer $\mathbf{A}$ in an autoregressive manner. It closely follows the prefix tuning technique for prompting a language model to produce an output of a particular style \cite{li2021prefix}, such as in our case an answer given a question and an image\footnote{Code available at: \url{github.com/tjvsonsbeek/open-ended-medical-vqa}}.

\textbf{Vision encoding stream} For encoding the image, we employ a pre-trained vision encoder to extract visual features $\{x_1, x_2 ... x_{\ell_x}\}$. To use these features as input to the decoder, they should be mapped into the latent space of the language decoder.
Following \cite{mokady2021clipcap}, we define a mapping network $\mathbf{f_M}$, implemented as a three-layer MLP. This network maps the visual features into a visual prefix $\{v_1, v_2, \dots v_x\} \in \mathbb{R}^{\ell_x \times e}$ for the language model, where $e$ is the embedding size. 

\textbf{Language encoding stream} Regarding the encoding of the textual part, firstly we utilize a standard tokenization process to obtain a sequence of tokens, both for the question $\mathbf{Q} = \{q_1, q_2 ... q_{\ell_q}\}\in \mathbb{R}^{\ell_q \times e}$ and answer $\mathbf{A} = \{a_1, a_2 ... a_{\ell_a}\}\in \mathbb{R}^{\ell_a \times e}$. This is followed by embedding the tokens using the embedding function of a pre-trained language model.

\textbf{Prompt structure}  To create a structured prompt, following existing QA methods using language models \cite{brown2020language,radford2019language}, we prepend the question, image, and answer tokens with tokenized descriptive strings, namely \texttt{question:}, \texttt{context:} and \texttt{answer:}. By placing the embeddings of the question before the visual tokens we mitigate the problem of fixation of the language model on the question \cite{luo2022biogpt,mokady2021clipcap}. As an example this would yield the following prompt template: $p=$\texttt{[question: What does the right side of the field show? context:  $v_1, v_2, \dots v_x$ answer: ]} which is fed as input to the language model.

\textbf{Language model} 
Following standard language modeling systems, we treat VQA as a conditional generation of text, and we optimize the standard maximum likelihood objective during training.
The language model receives the prompt sequence $p$ as input and outputs the answer $\mathbf{A}$, token by token. Specifically, at each time step $i$, the output of the model are the logits parametrizing a categorical distribution $p_\theta(\mathbf{A})$ over the vocabulary tokens. This distribution is represented as follows:
\begin{equation}
    \log p_\theta(\mathbf{A}) = \sum_{l_a}\log\:p_\theta(a_i|q_1, ... q_{\ell_q}, v_1, ... v_x, a_1,...a_{i-1}).
\end{equation}

The parameters of the language model are initialized from a pre-trained model, which has been previously pre-trained on huge web-collected datasets. 

\subsection{Parameter-efficient Strategies for Fine-Tuning the Language Model}
\begin{figure}[!t]
    \centering
    \includegraphics[width=\linewidth]{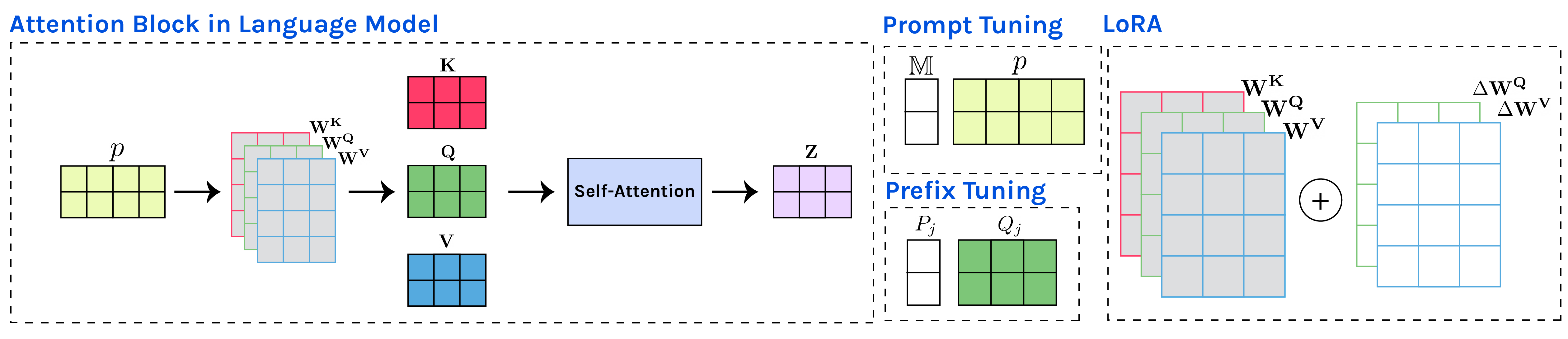}
    \caption{Parameter-efficient language model fine-tuning strategies used in our method.}
    \label{fig:tuning_LM}
\end{figure}

Standard fine-tuning of language models can hurt the generalization capabilities of the model, especially if small, domain-specific datasets are used as in our case. Therefore, we consider four different parameter-efficient strategies that adapt the attention blocks of language models, as illustrated in Fig.~\ref{fig:tuning_LM} and outlined below:

\textbf{Frozen method}: the parameters of the language model are kept entirely frozen during training, following~\cite{tsimpoukelli2021multimodal}. In this setting, only the mapping network is updated through backpropagation. \textbf{Prompt Tuning}: we prepend a set of $m$ learnable tokens $\mathbb{M}\in\mathbb{R}^{m\times e}$ to the input prompt sequence, which yields $[\mathbb{M},p]$ \cite{lester2021power} as input to the frozen language model. Besides updating the mapping network, this approach also involves updating these learnable tokens through backpropagation. \textbf{Prefix Tuning}: we prepend a learnable prefix $P_j$ to the query $Q_j$ of each attention block $j$ in the Transformer, such that $Q^{ft}_{j} = [P_j,Q_j]$ \cite{li2021prefix}. Similar as in prompt tuning, we update both the mapping function and the learnable prefixes of the queries. \textbf{Low-Rank Adaptation (LoRA)}: We add learnable weight matrices to the query $Q$ and value $V$ of the attention blocks in each layer of the frozen language model as $\mathbf{W}+\Delta\mathbf{W}$ following ~\cite{hu2021lora}. Again, the mapping function is trained together with the learnable weight matrices.

\section{Experimental Setup}
\subsubsection{Datasets}
The three datasets used for the evaluation of our method are Slake~\cite{liu2021slake}, PathVQA~\cite{he2020pathvqa}, and OVQA~\cite{huang2022ovqa}. These three datasets are the current most suitable VQA datasets given their large variety in answers and the manual curation of answers by domain experts. Each dataset is split 50/50 between `yes/no' and open-set answers. See the datatset details in Table~\ref{tab:datasets}. We use the official train/validation/test splits across all three datasets.
\begin{table}[!t]
\centering
\caption{Statistics of the medical VQA datasets used in this paper.}
\label{tab:datasets}
\setlength\tabcolsep{1.8em}
\resizebox{0.9\textwidth}{!}{%
\begin{tabular}{lccc}

\toprule
 & Slake & OVQA & PathVQA \\
\cmidrule(lr){2-2}\cmidrule(lr){3-3}\cmidrule(l){4-4} Number of images & 642& 2,001 & 4,998  \\
Number of questions  &14,028& 19,020 & 32,799 \\
Mean length of questions & $4.52$& $8.98$  & $6.36$  \\
Mean length of answers & $1.21$& $3.31$ & $1.80$  \\
Number of unique answers & 461& 641 & 3,182  \\

\bottomrule
\end{tabular}}
\end{table}

\subsubsection{Evaluation protocol}
We evaluate our approach using the conventional metrics BLEU-1 and F1 Score. Additionally, we measure the contextual capturing of information with BERTScore~\cite{zhangbertscore} this method can handle synonyms. Lastly to allow for comparison against existing classification-based methods we also report accuracy and F1 score.

\subsubsection{Implementation details}
We extract the visual features using a pre-trained CLIP model with ViT backbone~\cite{radford2021learning}, having a dimensionality of 512. The MLP layers of the mapping network $\mathbf{f_M}$ have sizes $\{512,(\ell_{x}\cdot e)/2, \ell_{x}\cdot e\}$. The length of
$\ell_x$ is set at 8. The lengths $\ell_q$ and $\ell_a$ are dataset dependent and defined by the mean number of tokens in the train set plus three times its standard deviation. Zero padding is added to the right side of the sequence for batch-wise learning. 

We use the following language models: GPT2-XL~\cite{radford2019language}, a causal language model with 1.5B parameters trained on WebText~\cite{radford2019language}. BioMedLM \cite{biomedlm} and BioGPT~\cite{luo2022biogpt} are both GPT2-based models, pre-trained on PubMed and biomedical data from The Pile~\cite{gao2020pile}, with a size of 1.5B and 2.7B parameters, respectively. All models are able to train on a single NVIDIA RTX 2080ti GPU (average training time $\approx$ 3 hours). We use the AdamW optimizer with 600 warmup steps and a learning rate of 5e-3 and apply early stopping with a tolerance of 3. 
\section{Results}
\begin{table}[!b]
\centering
\setlength\tabcolsep{0.40em}
\caption{Performance across different language models and fine-tuning strategies, measured in BLEU1 (BL1), BERTScore (BS), F1 and accuracy. Params\% is the amount of trainable parameters in the language model. Our method using GPT2 in combination with LoRA yields the best performance across all datasets.}
\label{tab:all}
\resizebox{\textwidth}{!}{%
\begin{tabular}{llcccccccccccccc}
\toprule
                                & LM fine-tuning & LM size & Params\% & \multicolumn{4}{c}{Slake} & \multicolumn{4}{c}{OVQA} & \multicolumn{4}{c}{PathVQA}  \\
    &              &  &      & \multicolumn{1}{c}{BL1} & \multicolumn{1}{c}{BS} & \multicolumn{1}{c}{F1}   & \multicolumn{1}{c}{Acc.} & \multicolumn{1}{c}{BL1} & \multicolumn{1}{c}{BS} & \multicolumn{1}{c}{F1}   & \multicolumn{1}{c}{Acc.} & \multicolumn{1}{c}{BL1} & \multicolumn{1}{c}{BS} & \multicolumn{1}{c}{F1}   & \multicolumn{1}{c}{Acc.} \\\cmidrule{1-4}\cmidrule(lr){5-8}\cmidrule(lr){9-12}\cmidrule(l){13-16}
\multicolumn{3}{l}{MedFuseNet~\cite{sharma2021medfusenet}}    & & \multicolumn{1}{c}{-}  & \multicolumn{1}{c}{-}      & \multicolumn{1}{c}{-} & \multicolumn{1}{c}{-} & \multicolumn{1}{c}{-}  & \multicolumn{1}{c}{-}      & \multicolumn{1}{c}{-} & \multicolumn{1}{c}{-}&  \multicolumn{1}{c}{60.5}  & \multicolumn{1}{c}{-}      & \multicolumn{1}{c}{38.1} & \multicolumn{1}{c}{-} \\\cmidrule{1-4}\cmidrule(lr){5-8}\cmidrule(lr){9-12}\cmidrule(l){13-16}

\multirow{4}{*}{\begin{tabular}[c]{@{}l@{}}Ours w/\\BioGPT\end{tabular}}     & Frozen  & \multirow{4}{*}{1.5B} & 0\%         & 64.5 & 69.9 & 57.7 & 66.5 & 32.4 & 71.9 & 52.5 & 53.5& 36.9 & 57.6 & 31.0 & 45.3  \\
    & Prefix \cite{li2021prefix}    &  & 0.487\%                           & 58.1 & 74.1 & 54.1 & 67.4 & 37.9 & 65.0 & 46.1 & 53.2& 53.6 & 61.8 & 34.8 & 46.7  \\
    & Prompt \cite{lester2021power}   &  & 0.001\%                           & 44.2 & \textBF{75.6} & 47.6 & 53.7 & 47.5 & 62.9 & 34.6 & 50.3& 28.0 & 58.7 & \textBF{43.8} & 33.2  \\
        & LoRA \cite{hu2021lora}        &  & 0.311\%                                & \textBF{59.2} & 72.2 & \textBF{63.1} & \textBF{71.9} & \textBF{41.0} & \textBF{68.5} & \textBF{57.7} & \textBF{57.3}& \textBF{57.8} & \textBF{62.9} & 40.4 & \textBF{47.9}  \\\cmidrule{1-4}\cmidrule(lr){5-8}\cmidrule(lr){9-12}\cmidrule(l){13-16}
\multirow{4}{*}{\begin{tabular}[c]{@{}l@{}}Ours w/\\BioMedLM\end{tabular}} & Frozen & \multirow{4}{*}{2.7B}  & 0\%          & 70.2 & 77.8 & 47.8 & 66.0 & 55.2 & 72.9 & 54.2 & 61.1& 61.2 & 66.1 & 52.4 & 53.0  \\

    & Prefix \cite{li2021prefix}     &  & 0.753\%                           & 64.3 & 79.4 & 60.9 & 63.3 & 49.1 & 76.9 & 51.5 & 60.1& 59.7 & 60.7 & 48.9 & 52.3  \\
    & Prompt \cite{lester2021power}      &  & 0.009\%                           & 44.6 & 73.5 & 38.8 & 41.6 & 48.9 & 72.8 & 44.3 & 59.5& 51.9 & 59.8 & 38.9 & 49.3  \\
    & LoRA \cite{hu2021lora}        &  & 0.101\%                               & \textBF{72.3} & \textBF{80.6} & \textBF{62.4} & \textBF{71.7} & \textBF{59.0} & 76.2 & \textBF{62.6} & \textBF{67.8}& \textBF{67.9} & \textBF{76.0} & \textBF{54.4} & \textBF{57.2}  \\
    \cmidrule{1-4}\cmidrule(lr){5-8}\cmidrule(lr){9-12}\cmidrule(l){13-16}
\multirow{4}{*}{\begin{tabular}[c]{@{}l@{}}Ours w/\\GPT2\end{tabular}}       & Frozen & \multirow{4}{*}{1.5B}       & 0\%  & 65.1 & 83.3 & 57.7 & 71.2 & 60.2 & 79.8 & 59.4 & 66.1& 64.2 & 74.6 & 47.5 & 58.1  \\

    & Prefix \cite{li2021prefix}     &  & 0.492\%                      & 70.0 & 86.5 & 66.3 & 74.1 & 61.2 & 83.9 & 65.5 & 68.9& 67.5 & 76.2 & 52.5 & 60.5 \\
    & Prompt \cite{lester2021power}     &  & 0.003\%                      & 57.8 & 80.3 & 49.9 & 60.0 & 57.8 & 78.3 & 55.2 & 63.1& 54.4 & 72.0 & 38.1 & 46.6 \\\rowcolor{gray!15}
   & LoRA \cite{hu2021lora}       &  & 0.157\%                            & \textBF{78.6} & \textBF{91.2} & \textBF{78.1} & \textBF{83.3} & \textBF{61.8} & \textBF{85.4} & \textBF{69.1} & \textBF{71.0}& \textBF{70.3} & \textBF{78.5} & \textBF{58.4} & \textBF{63.6} \\
    \bottomrule
\end{tabular}%
}
\end{table}

\subsubsection{Benefits of parameter-efficient fine-tuning} The evaluation of our method across various language models and fine-tuning settings in Table~\ref{tab:all} shows that language models can perform open-ended medical VQA. Specifically, we outperform the only existing method MedFuseNet \cite{sharma2021medfusenet} that does open-ended VQA, due to the capability of pre-trained language models to capture long-term dependencies when generating free-form answers. Additionally, prefix \cite{li2021prefix} and prompt tuning \cite{lester2021power} do not improve the performance of the model as much as using LoRA \cite{hu2021lora} which directly adapts the $Q$ and $V$ weight matrices of the attention blocks. Moreover, larger datasets show the most consistent performance gain of parameter-efficient fine-tuning across all metrics.  
\subsubsection{Comparison between standard and medical LMs} 
Using a language model pre-trained on a general text corpus, such as GPT2~\cite{radford2019language}, improves the overall performance compared to its medically-trained models (e.g. BioGPT or BioMedLM), as can be observed in Table~\ref{tab:all}. BioGPT and BioMedLM could be overoptimized to their medical text corpora, which leads to lack of generalization to different downstream domains.

As mentioned in \cite{luo2022biogpt,biomedlm}, these models require full fine-tuning on the respective downstream tasks, to achieve the desired performance. On the other hand, GPT2 benefits from observing diverse data during pre-training which also encompasses medically oriented text. This enables GPT2 models to generalize easily to other domains, which is relevant for our different VQA datasets.

\begin{table}[!t]
\setlength\tabcolsep{0.75em}
\centering
\caption{Comparison of the accuracy between open-ended VQA against classification-based VQA methods, split between yes/no and open-set answers. Our method performs particularly well on both types of answers compared to the state-of-the-art methods.  }
\label{tab:comparison}
\resizebox{\textwidth}{!}{%
\begin{tabular}{lluccuccucc}
    \toprule
                &  & \multicolumn{3}{c}{Slake} & \multicolumn{3}{c}{OVQA} & \multicolumn{3}{c}{PathVQA} \\\cmidrule(lr){3-5}\cmidrule(lr){6-8}\cmidrule(l){9-11}
    &
    &
    \multicolumn{1}{c}{Open-set} &
    \multicolumn{1}{c}{Yes/no} &
    \multicolumn{1}{c}{All} &
    \multicolumn{1}{c}{Open-set} &
    \multicolumn{1}{c}{Yes/no} &
    \multicolumn{1}{c}{All} &
    \multicolumn{1}{c}{Open-set} &
    \multicolumn{1}{c}{Yes/no} &
    \multicolumn{1}{c}{All} \\\cmidrule(lr){3-5}\cmidrule(lr){6-8}\cmidrule(l){9-11}
    MEVF-SAN~\cite{nguyen2019overcoming}          && 75.3 & 78.4 & 76.5 & 36.9 & 72.8 & 58.5& 6.0  & 81.0 & 43.6  \\
    MEVF-BAN~\cite{nguyen2019overcoming}           && 77.8 & 79.8 & 78.6 & 36.3 & 76.3 & 60.4& 8.1  & 81.4 & 44.8  \\
    MEVF-SAN+VQAMix~\cite{gong2022vqamix} && -    & -    & -     & -    & -    & - & 12.1 & 84.4 & 48.4   \\
    MEVF-BAN+VQAMix~\cite{gong2022vqamix} && -    & -    & -     & -    & -    & - & 13.4 & 83.5 & 48.6   \\
    MMQ-SAN~\cite{do2021multiple}           && -    & -    & -    & 56.9 & 76.2 & 68.5& 9.6  & 83.7 & 46.8  \\
    MMQ-BAN~\cite{do2021multiple}            && -    & -    & -    & 48.2 & 76.2 & 65.0& 11.8 & 82.1 & 47.1  \\
    QCR-BAN~\cite{zhan2020medical}            && 78.8 & 82.0& 80.0   & 52.6 & 77.7 & 67.7& -    & -    & -    \\
    CRPD-BAN~\cite{liu2021contrastive}          && 81.2 & 84.4 & 82.1 & -    & -    & -    & -    & -    & -    \\
    MMBERT~\cite{khare2021mmbert}           && -    & -    & -    & 37.9 & 80.2 & 63.3& -    & -    & -    \\
    QCR-CLIP~\cite{eslami2021does}           && 78.4    & 82.5    & 80.1    & -&-&-& -    & -    & -    \\\midrule\midrule
Ours w/ BioGPT (LoRA)   &  & 71.1    & 72.7   & 71.9       &  48.3  & 66.5   & 57.3& 30.2    & 65.5    & 47.9   \\
Ours w/ BioMedLM (LoRA) &  & 72.1    & 71.4   & 71.7   &  55.3  & 80.3   & 67.8& 34.1    & 80.4    & 57.2       \\
Ours w/ GPT2 (LoRA)     &  & \textBF{84.3}    & 82.1   & \textBF{83.3}   & \textBF{62.6}    & \textBF{84.7}    & \textBF{71.0}& \textBF{40.0}   & \textBF{87.0}   & \textBF{63.6}    \\
\bottomrule
\end{tabular}%
}
\end{table}

\subsubsection{Benefit of open-ended answer generation} Our method is performing significantly better on the open-set answering, in comparison to classification-based methods, as shown in Table~\ref{tab:comparison}. We also confirm that CLIP based image embeddings perform well in the medical domain~\cite{eslami2021does} compared to the conventional use of CNNs. Since our approach is generative, it is not bounded by the class imbalance issue, which is considered a bottleneck of classification-based VQA models. Our method performs especially well compared to other method on PathVQA, which relatively has the largest class imbalance, accentuating this effect. Even on the simple `yes/no' questions, the performance is better, showing that this simple yet effective method provides a more natural way of doing VQA.  

It worth noting that the comparison of accuracy as a metric for exact matches, between classification and generation methods is not in favor of generative methods. Despite that, we outperform existing methods on all datasets and metrics, which is a testament to the benefit of phrasing VQA as an open-ended generation problem. 

\begin{figure}[!t]
    \centering
    \includegraphics[width=\linewidth]{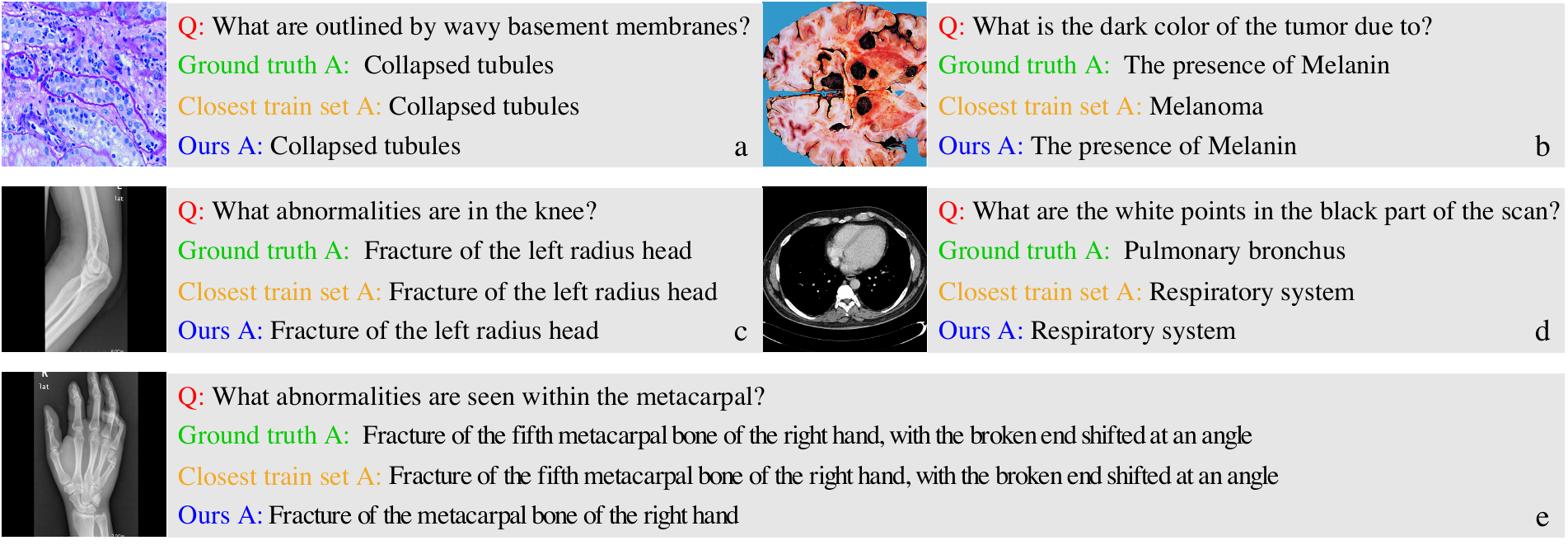}
    \caption{Outputs of our open-ended VQA method, with GPT2 and LoRA fine-tuning, using data samples from PathVQA~(a, b, d) and OVQA~(c, e).}
    \label{fig:eval}
\end{figure}

In Fig~\ref{fig:eval}(a-c), we show qualitative examples of capability of the language model to successfully predict the correct answer. However, in Fig~\ref{fig:eval}~(d, e) we show cases where our method predicts a factually correct answer which is not specific enough.


\begin{table}[!t]
\setlength\tabcolsep{0.70em}
\centering
\caption{Effect of using different prompt structures. Note that $\mathbf{Q}$ and $\mathbf{I}$ denote the question and image respectively. The regular setting with the question embeddings followed by the visual prefix (Fig.~\ref{fig:model_arch}) leads to the best overall performance. }
\label{tab:abl}
\resizebox{\textwidth}{!}{%
\begin{tabular}{lcccccccccccc}
\toprule
                                \multirow{2}{*}{Setting} & \multicolumn{4}{c}{Slake} & \multicolumn{4}{c}{OVQA} & \multicolumn{4}{c}{PathVQA}  \\\cmidrule(lr){2-5}\cmidrule(lr){6-9}\cmidrule(l){10-13}
          & \multicolumn{1}{c}{ B1 }&\multicolumn{1}{c}{BS} & \multicolumn{1}{c}{F1}   & \multicolumn{1}{c}{Acc.} & \multicolumn{1}{c}{ B1 } & \multicolumn{1}{c}{BS} & \multicolumn{1}{c}{F1}   & \multicolumn{1}{c}{Acc.} & \multicolumn{1}{c}{ B1 } & \multicolumn{1}{c}{BS} & \multicolumn{1}{c}{F1}   & \multicolumn{1}{c}{Acc.} \\\midrule

w/o $\mathbf{Q}$&29.4 & 48.4 & 14.3 & 22.1 & 33.2 & 41.9 & 18.1 & 27.6 & 42.9 & 43.8 & 18.3 & 24.6 \\
w/o $\mathbf{I}$\:&54.8 & 79.3 & 49.5 & 50.9 & 45.5 & 77.1 & 49.5 & 54.4 & 65.9 & 72.8 & 47.2 & 46.3 \\
Swap $\mathbf{Q}$ and $\mathbf{I}$&73.3 & 88.7 & 73.2 & 74.9 & 60.0   & 84.2 & 67.3 & 66.9 & 70.2 & 78.0  & 57.2 & 58.7\\\cmidrule(r){1-1}\cmidrule(lr){2-5}\cmidrule(lr){6-9}\cmidrule(l){10-13}
Regular& \textBF{78.6} & \textBF{91.2} & \textBF{78.1} & \textBF{83.3} & \textBF{61.8} & \textBF{85.4} & \textBF{69.1} & \textBF{71.0}& \textBF{70.3} & \textBF{78.5} & \textBF{58.4} & \textBF{63.6} \\\bottomrule
\end{tabular}
}
\end{table}

\subsubsection{Effect of using different prompt structures}
We also investigate the influence of the prompt structure on the overall performance, demonstrated in Table \ref{tab:abl}. It can be observed that the performance largely decreases when the question is removed, compared to when the visual information is removed. This suggests that the question plays a more important role in answer generation.
Interestingly, the model is sensitive the order of the elements in the prompt, as the swapping of the question embeddings and the visual prefix yields decreases the performance. The reason for this is that the language model conveys lower to no importance the visual information if it is located in front of the question. In this situation the language model basically generates blind answers. This highlights the importance of prompt structure.

\section{Conclusion}

In this paper, we propose a new perspective on medical VQA. We are using generative language models to generate answers in an open-ended manner, instead of performing a closed-set classification. Additionally, by using various parameter-efficient fine-tuning strategies we are able to use language models with billions of parameters, even though dataset sizes in this domain are small. This leads to excellent performance compared to classification-based methods. In conclusion, our approach offers a more accurate and efficient solution for medical VQA. 

\section*{Acknowledgements}
This work is financially supported by the Inception Institute of Artificial Intelligence, the University of Amsterdam and the allowance Top consortia for Knowledge and Innovation (TKIs) from the Netherlands Ministry of Economic Affairs and Climate Policy.

\bibliographystyle{splncs04}
\bibliography{sources.bib}
\end{document}